\documentclass[11pt]{article}
\usepackage[margin=1in]{geometry}
\usepackage{times}
\usepackage{amsmath,amssymb}
\usepackage{graphicx}
\usepackage{booktabs}
\usepackage{multirow}
\usepackage{hyperref}
\usepackage{caption}
\usepackage{subcaption}
\usepackage[numbers,sort&compress]{natbib}
\usepackage{xcolor}
\usepackage{float}
\usepackage{microtype}
\usepackage{xurl}

\hypersetup{colorlinks=true, linkcolor=blue, citecolor=blue, urlcolor=blue}

\title{Parser, Chunking, and Embedding Interactions in Retrieval-Augmented Generation over Indian Government Regulatory Documents}

\author{Shubham Kumar Singh}
\date{September 2026}

\begin{document}
\maketitle

\begin{abstract}
Retrieval-augmented generation (RAG) pipelines are typically assembled from independently-chosen components -- a document parser, a chunking strategy, and an embedding model -- yet these choices are rarely evaluated jointly, and evaluations that do combine them are usually run on a single document or corpus. We present a controlled factorial study of $3$ parsers $\times$ $3$ chunking strategies $\times$ $5$ dense embedding models, together with a sparse (BM25) baseline, evaluated against $800$ question instances, each with one or more required evidence strings, with evidence strings automatically validated against source text and a $10\%$ random sample manually reviewed, spread across four structurally distinct Indian central-government regulatory documents (leave rules, conduct rules, a large rule compilation, and a right-to-information statute). We fit linear mixed-effects models with document-query-level random intercepts to the resulting $72{,}000$-row result set, run Holm-corrected paired comparisons between matched dense and sparse configurations, and report clustered bootstrap confidence intervals for all $54$ unique retriever configurations. We find that (i) no single retriever family dominates across documents -- BM25 significantly outperforms dense retrieval on some documents while the reverse holds on others; (ii) parser and chunker choice interact significantly, and the best-performing combination is not predictable from either component's individual ranking; (iii) one widely-used general-purpose embedding model (MPNet-base) is a significant, consistent underperformer, with a severe and specific failure mode on table-derived questions; and (iv) our corpus exhibits a near-saturated evidence-preservation ceiling ($>98\%$) across all parsing/chunking combinations, meaning that the retrieval differences we measure are driven almost entirely by ranking quality rather than information loss during ingestion. We additionally report embedding-dimension and chunk-size/overlap ablations and an efficiency/quality Pareto analysis. We release our full evaluation harness, corpus manifest, and 800-question benchmark.
\end{abstract}

\section{Introduction}

Retrieval-augmented generation (RAG) systems built over enterprise, legal, or regulatory documents are assembled from a pipeline of independent design choices: how the source document is parsed into text elements, how those elements are chunked for indexing, and which embedding model is used to represent chunks for dense retrieval. Practitioners frequently select these components independently -- picking a popular embedding model off a leaderboard, and a chunking strategy from a framework's default configuration -- without testing whether the choices interact.

A small number of recent studies have begun to evaluate these components jointly. \citet{turkish2026} run a $3\times5\times2$ chunking $\times$ embedding $\times$ generator factorial design across three Turkish-language documents, producing $9{,}000$ graded QA evaluations with paired McNemar testing and Holm correction, and find that the best individual components do not necessarily compose into the best complete pipeline. \citet{hicbench2026} argue that many existing RAG benchmarks under-specify the relationship between evidence and chunk boundaries (\emph{evidence sparsity}), and introduce evidence-dense, multi-level chunking annotations to address it. Retrieval benchmarking more broadly is anchored by BEIR \citep{beir2021}, which established that no single dense retriever dominates across heterogeneous domains and that sparse baselines (BM25) remain highly competitive.

We extend this line of work along two axes. First, we hold the corpus \emph{domain} constant (statutory and regulatory text) while varying document \emph{structure} deliberately -- one table-heavy numerical-schedule document, one prescriptive prohibition/exception document, one large heterogeneous multi-chapter compilation, and one long-form definitional legal text -- so that interactions between parser/chunker choice and document structure can be examined while reducing uncontrolled variation in domain and language. Second, we evaluate the full $3\times3\times5$ parser $\times$ chunker $\times$ embedding grid against a sparse baseline on the same $800$-question benchmark, and report both aggregate leaderboards and question-level paired statistics, rather than aggregate metrics alone.

\paragraph{Research questions.} We organize our analysis around four questions:
\begin{itemize}
    \item[\textbf{RQ1}] Does parser choice (raw-text extraction vs.\ heading-aware extraction vs.\ a layout-aware document-conversion tool) affect retrieval quality, and does this effect depend on the chunking strategy applied afterward?
    \item[\textbf{RQ2}] Does any single dense embedding model dominate across documents of different structure, and how does the best dense configuration compare to a sparse (BM25) baseline?
    \item[\textbf{RQ3}] How much of the measured retrieval quality is attributable to information loss during parsing/chunking (a \emph{representation ceiling}) as opposed to ranking quality at retrieval time?
    \item[\textbf{RQ4}] How sensitive is retrieval quality to two common practical levers -- embedding dimensionality and chunk size/overlap -- and what does this imply for deployment cost trade-offs?
\end{itemize}

\paragraph{Contributions.}
\begin{itemize}
    \item An $800$-question, four-document, structurally-stratified benchmark for regulatory/statutory RAG, with per-question evidence strings validated against source text and manually spot-checked (10\% random sample, with duplicate removal against the corpus authors' own held-out queries).
    \item A full $3\times3\times5$ parser $\times$ chunker $\times$ embedding factorial evaluation plus a BM25 baseline, fit with mixed-effects models and reported with clustered bootstrap confidence intervals and Holm-corrected paired tests.
    \item A corpus-level \emph{evidence preservation audit} that separately quantifies how much gold evidence survives parsing/chunking versus how much is actually retrieved, isolating representation loss from ranking failure.
    \item Embedding-dimension and chunk-size/overlap ablations, run under configuration choices fixed prior to each sweep (a pre-declared parser/chunker selection rule for the dimension ablation; a fixed parser/embedding pair for the chunk-size/overlap sweep) to avoid post-hoc cherry-picking, plus a latency/quality Pareto analysis.
\end{itemize}

\section{Related Work}

\paragraph{RAG evaluation.} RAGAS \citep{ragas2023} and RAGChecker \citep{ragchecker2024} introduced fine-grained metrics separating retrieval/context quality from generation quality, but typically evaluate a single fixed retrieval configuration. Our study is retrieval-only by design: we deliberately exclude a generation stage so that the effect of parser/chunker/embedding choice on retrieval is not confounded with generator behavior.

\paragraph{Retrieval benchmarking.} BEIR \citep{beir2021} remains the standard reference for cross-domain retrieval evaluation and established that BM25 is a strong, sometimes-dominant baseline outside of narrow in-domain settings -- a finding we replicate at the level of individual documents within a single narrow domain. T\textsuperscript{2}-RAGBench \citep{t2ragbench2026} similarly finds hybrid sparse+dense retrieval highly effective specifically for text-and-table content, consistent with our finding that dense embeddings are least reliable on table-derived question types.

\paragraph{Chunking evaluation.} HiChunk/HiCBench \citep{hicbench2026} motivate evidence-aware, multi-level chunking annotation. We adopt a related but simpler evidence model: each question is annotated with one or more required evidence strings, and we compute an \emph{evidence recall} metric per top-$k$ retrieved set (\S\ref{sec:metrics}) rather than committing to a single gold chunk, which avoids penalizing correct retrieval when a chunker's boundary decisions differ from a hand-authored gold segmentation.

\paragraph{Factorial pipeline evaluation.} The closest prior work is \citet{turkish2026}, which runs a comparable chunking $\times$ embedding $\times$ generator factorial design with paired significance testing. Our design differs in scope (parser $\times$ chunker $\times$ embedding, no generation stage) and in domain (English-language Indian regulatory/statutory text vs.\ Turkish-language documents), and we additionally report a corpus-level evidence-preservation ceiling that separates representation loss from retrieval failure, which that work does not isolate.

\section{Corpus Construction}
\label{sec:corpus}

\subsection{Source documents}

We select four publicly available, freely redistributable Government of India regulatory and statutory documents, chosen to be structurally heterogeneous while remaining within a single administrative/legal domain (Table~\ref{tab:corpus}). All four documents are official rules or Acts intended for public reference and carry no confidentiality restriction.

\begin{table}[H]
\centering
\small
\begin{tabular}{@{}lllp{4.7cm}@{}}
\toprule
ID & Document & Issuing authority & Dominant structural character \\
\midrule
D1 & Central Civil Services (Leave) Rules, 1972 & Dept.\ of Personnel \& Training & Numerical entitlement tables, accrual/encashment schedules \\
D2 & CCS (Conduct) Rules, 1964 & Dept.\ of Personnel \& Training & Prescriptive prohibitions, exceptions, disciplinary language \\
D3 & Fundamental Rules \& Supplementary Rules, Part I & Dept.\ of Personnel \& Training & Large, heterogeneous, multi-chapter rule hierarchy \\
D4 & Right to Information Act, 2005 & Parliament of India & Long-form legal narrative, definitions, cross-referenced sections \\
\bottomrule
\end{tabular}
\caption{Corpus documents. Full source URLs and retrieval dates are recorded in the released corpus manifest with SHA-256 hashes of the exact PDF bytes used.}
\label{tab:corpus}
\end{table}

\subsection{Question--evidence generation}

For each document, source elements (paragraphs and, where present, table rows) were used to prompt an LLM under a fixed schema requiring: (i) a natural-language question grounded in exactly one (or, for cross-referencing questions, two) source element(s); (ii) a categorical \texttt{question\_type} label (e.g.\ \texttt{exact\_fact}, \texttt{table\_lookup}, \texttt{negation}, \texttt{procedural}, \texttt{numerical}, \texttt{comparison}); (iii) a \texttt{difficulty} label; and (iv) one or more \texttt{required\_evidence} strings that the model was instructed to copy verbatim from the source element rather than paraphrase. Candidate questions whose evidence strings could not be matched against the source document were discarded automatically. The surviving pool was reduced to $200$ questions per document ($800$ total), then \textbf{manually reviewed by the authors on a random 10\% sample} and checked for duplication against an independently-authored held-out query set, with duplicates removed.

We do not claim this process eliminates all annotation noise; we report it plainly as a limitation in \S\ref{sec:limitations} rather than presenting the benchmark as exhaustively human-authored.

\subsection{Evidence matching}

A retrieved chunk is considered to satisfy a required-evidence string if the (whitespace-normalized, lowercased) string appears as a substring of the chunk, \emph{or} if at least $80\%$ of the evidence string's tokens are present in the chunk. The $80\%$ threshold accommodates minor tokenization/whitespace divergence introduced by different parsers while limiting false matches caused by minor parser-induced tokenization differences.

\section{Experimental Design}
\label{sec:design}

\subsection{Parsers}
\begin{itemize}
    \item \textbf{normal}: raw per-page text extraction (via \texttt{pypdf}), split on blank-line boundaries into elements.
    \item \textbf{context\_aware}: as \texttt{normal}, but paragraphs are classified as headings via a rule-based pattern (numbered/lettered section markers, all-caps short lines, Markdown-style headers) and each subsequent body element is prefixed with its most recent heading as a section tag.
    \item \textbf{docling}: layout-aware document conversion \citep{docling2025} to Markdown, split into blocks on blank lines.
\end{itemize}

\subsection{Chunkers}
\begin{itemize}
    \item \textbf{basic}: fixed-size token windows (256 tokens, no overlap) over the concatenated element stream.
    \item \textbf{medium}: recursive/sentence-boundary-aware chunking (256-token target, 48-token tail overlap carried from the previous chunk).
    \item \textbf{advanced}: hierarchical, section-aware chunking. Elements are first grouped into sections using the same heading heuristic as the \texttt{context\_aware} parser; each section is chunked independently (256 tokens, 32-token overlap) and every resulting chunk is prefixed with a \texttt{[Doc: \{id\} | Section: \{heading\}]} breadcrumb.
\end{itemize}
Note that \texttt{advanced} chunking performs its own section detection independently of parser choice, so its interaction with the \texttt{context\_aware} parser is not redundant by construction but is an empirical question we test directly (\S\ref{sec:interaction}).

\subsection{Retrievers}
Five sentence-embedding models spanning two size tiers and three training regimes: \texttt{all-MiniLM-L6-v2}, \texttt{all-mpnet-base-v2}, \texttt{bge-small-en-v1.5}, \texttt{bge-base-en-v1.5}, and \texttt{e5-base-v2}, each with model-appropriate query/passage prefixes, indexed with exact inner-product search (FAISS \texttt{IndexFlatIP}) over normalized embeddings. We additionally run \textbf{BM25Okapi} as a sparse baseline over the same chunk sets, with no query expansion or tuning (``BM25-Naive'').

\subsection{Metrics}
\label{sec:metrics}
For a query with $m$ required evidence strings and retrieved chunk list $c_1,\dots,c_k$, we compute:
\begin{itemize}
    \item \textbf{Evidence Recall@$k$}: the fraction of the $m$ evidence strings matched by the union of the top-$k$ chunks.
    \item \textbf{Hit@$k$} and \textbf{Full@$k$}: whether at least one, or all, evidence strings are matched within the top-$k$.
    \item \textbf{MRR}: reciprocal rank of the first chunk matching any evidence string.
    \item \textbf{nDCG@5, nDCG@10}: normalized discounted cumulative gain, using newly matched evidence as the per-rank gain.
    \item \textbf{Evidence Coverage}: the fraction of required evidence strings present in at least one chunk of the indexed corpus, regardless of retrieval rank, reported separately for corpus-level auditing (\S\ref{sec:ceiling}).
\end{itemize}
$k \in \{1,3,5,10\}$ throughout; our primary outcome variable for statistical modeling is Evidence Recall@10 (denoted \texttt{recall\_10}).

\subsection{Statistical methodology}
\label{sec:stats}
We fit two linear mixed-effects models on \texttt{recall\_10}, with a document-query-level random intercept to account for repeated measurement of the same question across configurations:
\begin{align}
\text{recall\_10} &\sim \text{Parser} * \text{Chunker} * \text{Embedding} + (1\mid \text{document:query}) \quad \text{[dense, } n{=}45\text{ cells]}\\
\text{recall\_10} &\sim \text{Parser} * \text{Chunker} + (1\mid \text{document:query}) \quad \text{[BM25, } n{=}9\text{ cells]}
\end{align}
For every (document, parser, chunker, embedding) cell we additionally run a paired two-sided Wilcoxon signed-rank test between the dense configuration and the query-matched BM25-Naive configuration on the same (parser, chunker, document), and apply Holm's step-down correction across all resulting comparisons. We report clustered bootstrap 95\% confidence intervals (2{,}000 resamples over query-level means) for all $54$ unique (parser, chunker, retriever) configurations.

\subsection{Pre-declared ablation selection rule}
To avoid post-hoc cherry-picking when selecting a fixed configuration for the \emph{embedding-dimension} ablation, we pre-declare the selection rule: \emph{the parser and chunker that jointly maximize mean dense Evidence Recall@5 in the main experiment}. This selected \texttt{normal} (parser) / \texttt{advanced} (chunker); the dimension ablation accordingly fixes this parser/chunker pair and varies only the embedding model, using a Matryoshka-style truncatable model (\texttt{nomic-embed-text-v1.5}) so that dimensionality can be varied post-hoc from a single 768-dimensional encoding without retraining, with truncation-and-renormalization mechanics validated before use (shape, unit-norm, and finite-score checks on all tested dimensions).

The \emph{chunk-size/overlap} ablation is a separate exercise with a different target: it varies the internal size and overlap hyperparameters of a single chunking \emph{algorithm} (\texttt{chunk\_recursive}, the mechanism underlying the \texttt{medium} chunker) rather than comparing across chunking strategies. Because the \texttt{advanced} chunker does not expose a single size/overlap parameter to sweep (it chunks per detected section rather than by fixed token windows), it is not a candidate for this ablation, so the pre-declared parser/chunker selection rule above does not apply here. We instead fix this ablation to the \texttt{context\_aware} parser and the \texttt{BGE-base} embedding model -- a configuration chosen prior to running the sweep and held constant throughout it -- while sweeping chunk size and overlap on \texttt{chunk\_recursive} directly.

\section{Results}

\subsection{Main leaderboard (RQ1, RQ2)}
Table~\ref{tab:leaderboard} shows the top dense configurations and the full BM25 leaderboard by Evidence Recall@10. The best overall configuration is \texttt{normal} parsing with \texttt{advanced} (hierarchical) chunking and the \texttt{BGE-small} embedding (Evidence Recall@10 $=0.958$), narrowly ahead of the same parser/chunker pairing with \texttt{E5-base} and of \texttt{docling}+\texttt{advanced}+\texttt{E5-base}. Notably, the simplest parser (\texttt{normal}) paired with the most structure-aware chunker (\texttt{advanced}) outperforms the heading-injecting \texttt{context\_aware} parser in every top-10 dense configuration; \texttt{context\_aware} does not appear until much further down the ranking. We discuss this in \S\ref{sec:interaction}.

\begin{table}[H]
\centering
\small
\begin{tabular}{@{}llllll@{}}
\toprule
Parser & Chunker & Embedding & Recall@5 & Recall@10 & MRR \\
\midrule
normal & advanced & BGE-small & 0.909 & \textbf{0.958} & 0.767 \\
normal & advanced & E5-base & 0.904 & 0.949 & 0.771 \\
docling & advanced & E5-base & 0.910 & 0.947 & 0.787 \\
normal & advanced & BGE-base & 0.895 & 0.945 & 0.772 \\
normal & basic & E5-base & 0.872 & 0.937 & 0.728 \\
\midrule
\multicolumn{6}{l}{\emph{BM25-Naive (sparse baseline)}} \\
\midrule
docling & advanced & --- & 0.873 & 0.918 & 0.778 \\
normal & basic & --- & 0.872 & 0.912 & 0.779 \\
normal & advanced & --- & 0.878 & 0.910 & 0.786 \\
\bottomrule
\end{tabular}
\caption{Top-5 dense configurations and top-3 BM25 configurations by Evidence Recall@10, averaged over all four documents.}
\label{tab:leaderboard}
\end{table}

\subsection{Parser--Chunker Interaction}
\label{sec:interaction}
\begin{figure}[H]
\centering
\includegraphics[width=0.62\textwidth]{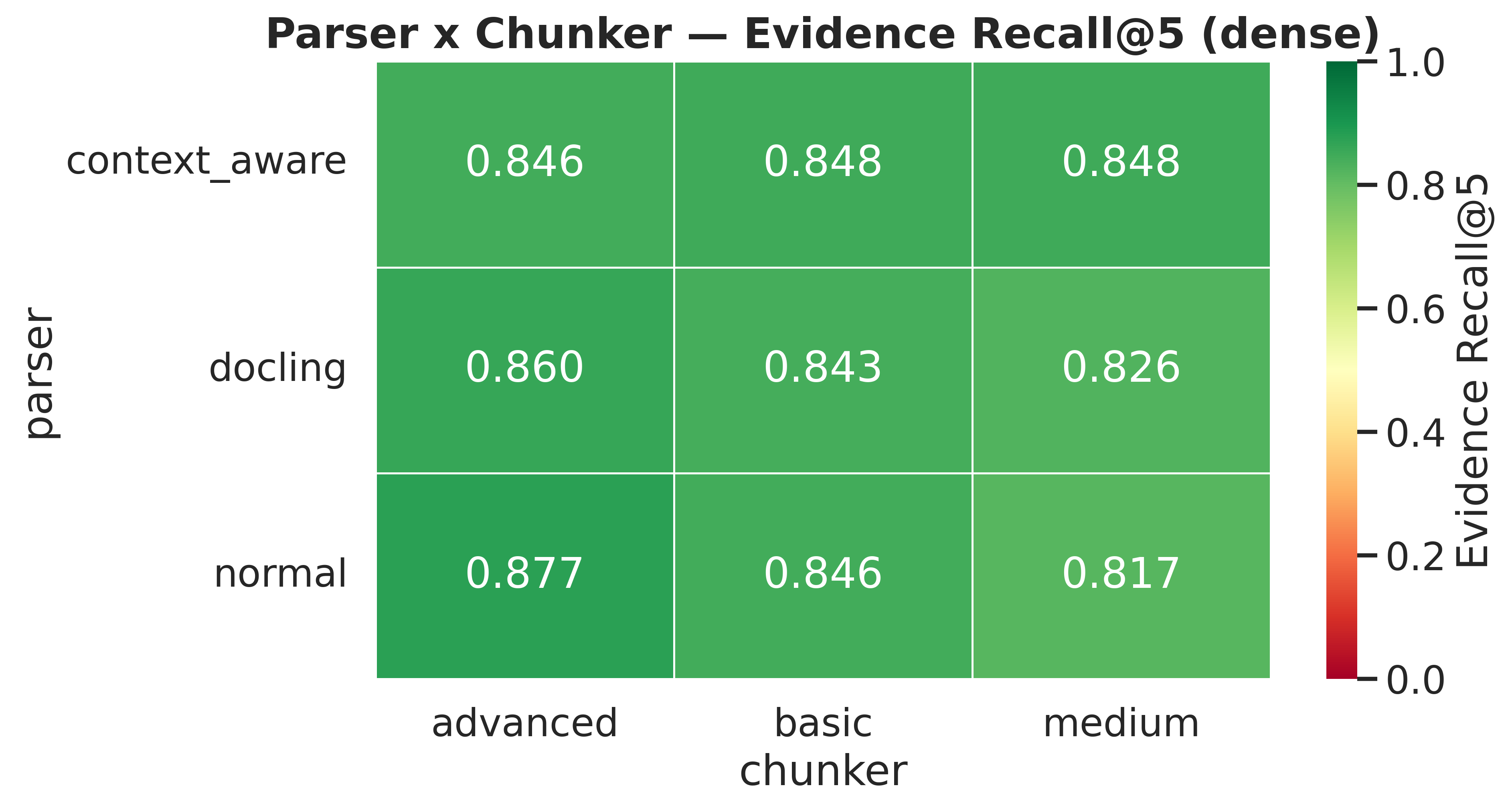}
\caption{Mean dense Evidence Recall@5 by parser and chunker, averaged over all five embedding models and four documents.}
\label{fig:parser_chunker}
\end{figure}

Figure~\ref{fig:parser_chunker} visualizes mean dense Evidence Recall@5 for every parser/chunker pair. The corresponding mixed-effects model on Evidence Recall@10 (Table~\ref{tab:mixedlm}) confirms a statistically significant parser $\times$ chunker interaction: relative to the \texttt{context\_aware}+\texttt{basic}+\texttt{BGE-base} reference cell, both \texttt{normal} ($\beta=0.039$, $p=0.0015$) and the \texttt{medium} chunker ($\beta=0.028$, $p=0.025$) carry significant positive main effects, but their interaction is significantly \emph{negative} ($\beta=-0.077$, $p<0.0001$) -- i.e.\ \texttt{normal}+\texttt{medium} performs \emph{worse} than either main effect alone would predict. \texttt{MPNet-base} carries a significant negative main effect independent of parser/chunker ($\beta=-0.049$, $p<0.0001$). The BM25 model shows a parallel pattern: both \texttt{docling} and \texttt{normal} parsers significantly outperform \texttt{context\_aware} ($p<0.0001$ each), but this advantage is significantly attenuated when paired with \texttt{basic} or \texttt{medium} chunking rather than \texttt{advanced}.

\begin{table}[H]
\centering
\small
\begin{tabular}{@{}lrrrl@{}}
\toprule
Term & Coef. & SE & $z$ & $p$ \\
\midrule
Intercept & 0.905 & 0.010 & 88.86 & $<0.0001$ \\
Parser[normal] & 0.039 & 0.012 & 3.17 & 0.0015 \\
Chunker[medium] & 0.028 & 0.012 & 2.25 & 0.0247 \\
Embedding[MPNet-base] & $-0.049$ & 0.012 & $-3.93$ & $<0.0001$ \\
Chunker[medium]:Embedding[MiniLM-L6] & $-0.035$ & 0.018 & $-1.97$ & 0.0493 \\
Parser[docling]:Chunker[medium] & $-0.047$ & 0.018 & $-2.69$ & 0.0073 \\
Parser[normal]:Chunker[medium] & $-0.077$ & 0.018 & $-4.39$ & $<0.0001$ \\
\bottomrule
\end{tabular}
\caption{Significant ($p<0.05$) fixed effects, dense $3\times3\times5$ mixed-effects model on Evidence Recall@10. Reference level: \texttt{context\_aware} / \texttt{basic} / \texttt{BGE-base}. Full table in the released results.}
\label{tab:mixedlm}
\end{table}

\subsection{Dense vs.\ sparse retrieval (RQ2)}
\label{sec:densevsbm25}
\begin{figure}[H]
\centering
\includegraphics[width=0.62\textwidth]{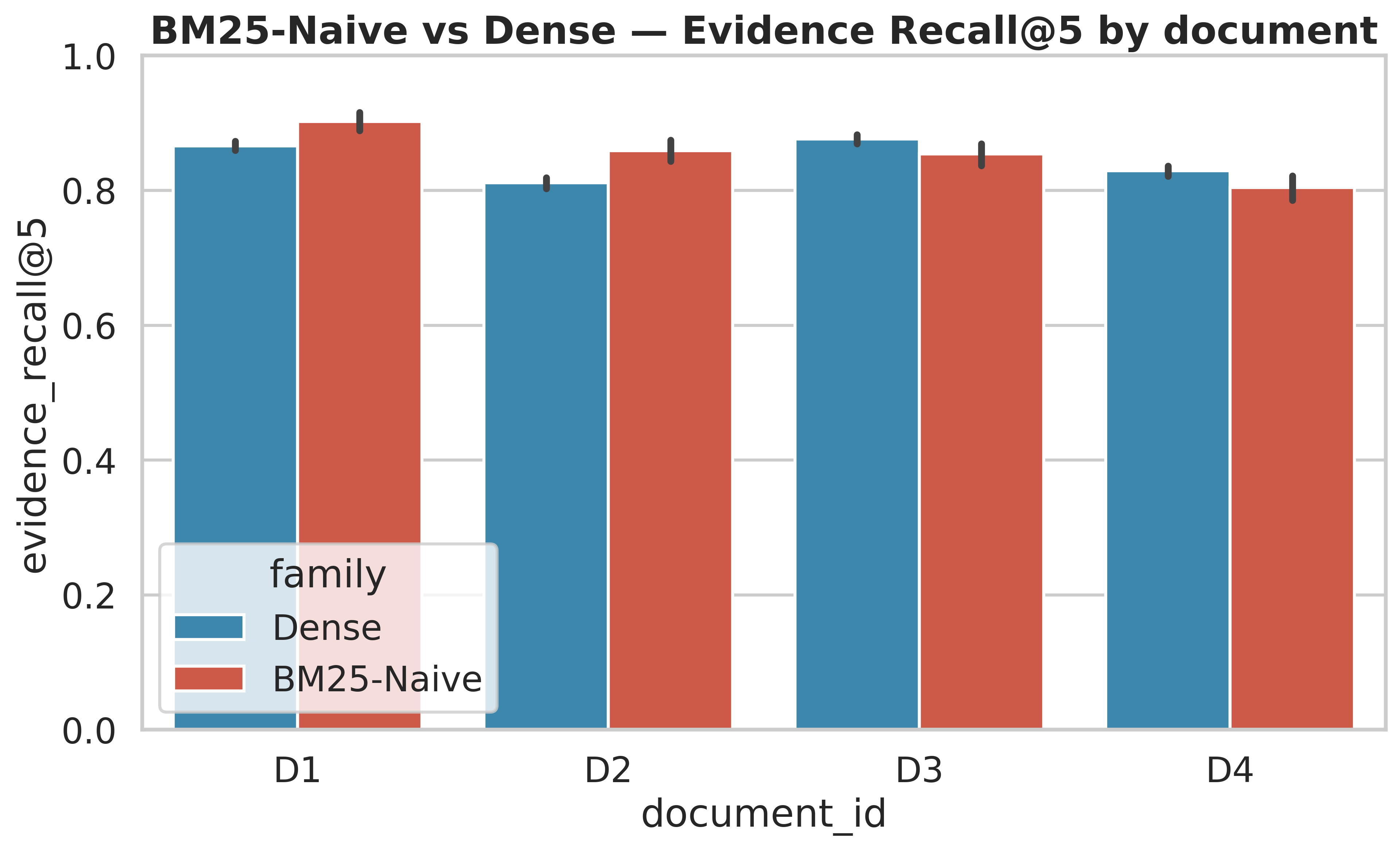}
\caption{Evidence Recall@5, dense (best-of-5, mean) vs.\ BM25-Naive, by document.}
\label{fig:dense_vs_bm25}
\end{figure}

Averaged across all documents, three of five embedding models (\texttt{E5-base}, \texttt{BGE-base}, \texttt{BGE-small}) beat matched BM25 in $72$--$86\%$ of parser/chunker cells (Table~\ref{tab:densevsbm25}), while \texttt{MPNet-base} loses to BM25 in $72\%$ of cells with a negative mean difference. Critically, this advantage is \emph{not} uniform across documents (Figure~\ref{fig:dense_vs_bm25}): BM25 wins on D1 and D2, dense wins on D3 and D4, and several of these per-cell differences remain significant after Holm correction across the full comparison set (e.g.\ D2/\texttt{docling}/\texttt{basic}/\texttt{MPNet-base}: BM25 $=0.92$ vs.\ dense $=0.71$, $p_{\text{Holm}}<0.0001$; D4/\texttt{context\_aware}/\texttt{advanced}/\texttt{E5-base}: BM25 $=0.87$ vs.\ dense $=0.96$, $p_{\text{Holm}}=0.021$). This directly answers RQ2: no retriever family dominates across our corpus, and the choice of sparse vs.\ dense should be validated per-document rather than assumed from an aggregate leaderboard.

\begin{table}[H]
\centering
\small
\begin{tabular}{@{}lrrr@{}}
\toprule
Embedding & Mean diff.\ vs.\ BM25 & \% cells won & \% significant wins (Holm) \\
\midrule
E5-base & $+0.036$ & 86.1\% & 2.8\% \\
BGE-base & $+0.029$ & 80.6\% & 2.8\% \\
BGE-small & $+0.024$ & 72.2\% & 0.0\% \\
MiniLM-L6 & $+0.003$ & 50.0\% & 0.0\% \\
MPNet-base & $-0.034$ & 27.8\% & 0.0\% \\
\bottomrule
\end{tabular}
\caption{Dense vs.\ matched BM25-Naive on Evidence Recall@10, averaged over all (document, parser, chunker) cells. ``\% significant wins'' is the fraction of individual cells where dense significantly beats BM25 after Holm correction across the full comparison family; the low rate reflects the conservativeness of Holm correction at $n{=}180$ comparisons ($4$ documents $\times$ $3$ parsers $\times$ $3$ chunkers $\times$ $5$ embeddings) with modest per-cell query counts, not an absence of a real aggregate effect.}
\label{tab:densevsbm25}
\end{table}

\subsection{Failure modes by question type}
\begin{figure}[H]
\centering
\includegraphics[width=0.78\textwidth]{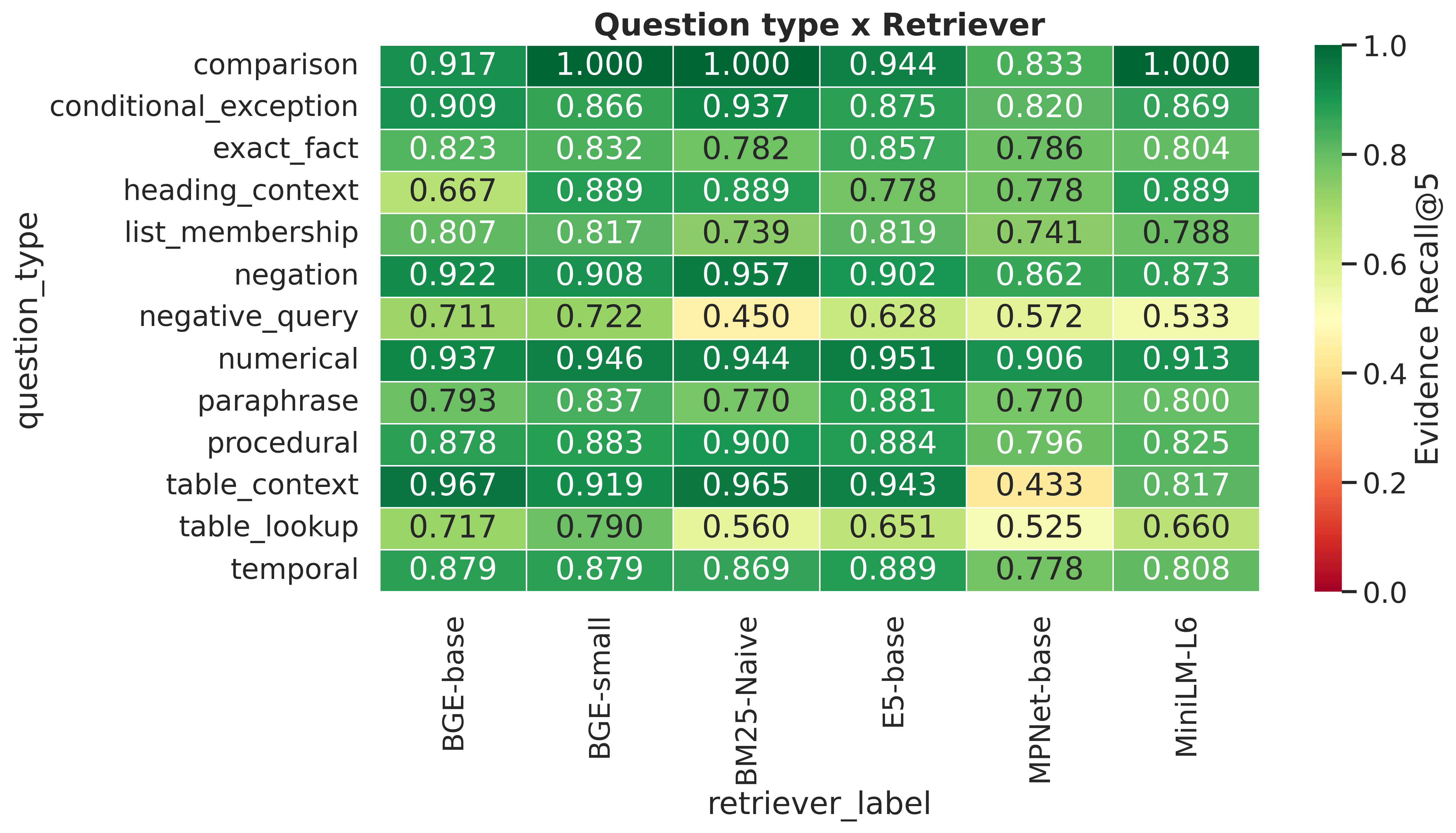}
\caption{Evidence Recall@5 by question type and retriever.}
\label{fig:question_type}
\end{figure}

Figure~\ref{fig:question_type} breaks down Evidence Recall@5 by question type and retriever. Two patterns stand out. First, \texttt{table\_lookup} and \texttt{negative\_query} questions are among the hardest categories for most retrievers (dense and sparse alike; e.g.\ BM25 scores $0.56$ and $0.45$ respectively, its two lowest categories), suggesting these failure modes are structural to the retrieval task rather than specific to any one model. Second, \texttt{MPNet-base} exhibits a severe, isolated failure on \texttt{table\_context} questions ($0.433$, compared to $0.79$--$0.97$ for the other retrievers on the same category) while performing only moderately worse than other models elsewhere -- a specific, reproducible weakness rather than uniformly poor performance.

\subsection{Representation ceiling audit (RQ3)}
\label{sec:ceiling}
We separately measure, for every (document, parser, chunker) combination, what fraction of gold evidence is present in \emph{some} chunk in the resulting corpus (regardless of whether it is ever retrieved). Mean coverage ranges narrowly from $0.983$ to $0.998$ across all nine parser/chunker combinations (Table~\ref{tab:ceiling}), and the fraction of queries with \emph{complete} evidence coverage ranges from $0.980$ to $0.996$. This indicates that the vast majority of the retrieval-quality variation reported in \S\ref{sec:interaction}--\ref{sec:densevsbm25} appears to arise from \emph{ranking} quality at retrieval time rather than information loss during parsing or chunking -- our parsers and chunkers all preserve the source text well; they differ in how retrievable that preserved text is.

\begin{table}[H]
\centering
\small
\begin{tabular}{@{}llrr@{}}
\toprule
Parser & Chunker & Mean coverage & \% queries fully covered \\
\midrule
normal & advanced & 0.998 & 99.6\% \\
docling & advanced & 0.995 & 99.1\% \\
normal & basic & 0.991 & 98.9\% \\
context\_aware & basic & 0.988 & 98.5\% \\
context\_aware & medium & 0.988 & 98.5\% \\
docling & medium & 0.987 & 98.3\% \\
docling & basic & 0.985 & 98.0\% \\
context\_aware & advanced & 0.984 & 98.3\% \\
normal & medium & 0.983 & 98.0\% \\
\bottomrule
\end{tabular}
\caption{Corpus-level evidence preservation (recall ceiling), averaged across all four documents.}
\label{tab:ceiling}
\end{table}

\subsection{Ablations (RQ4)}

\begin{figure}[H]
\centering
\includegraphics[width=0.85\textwidth]{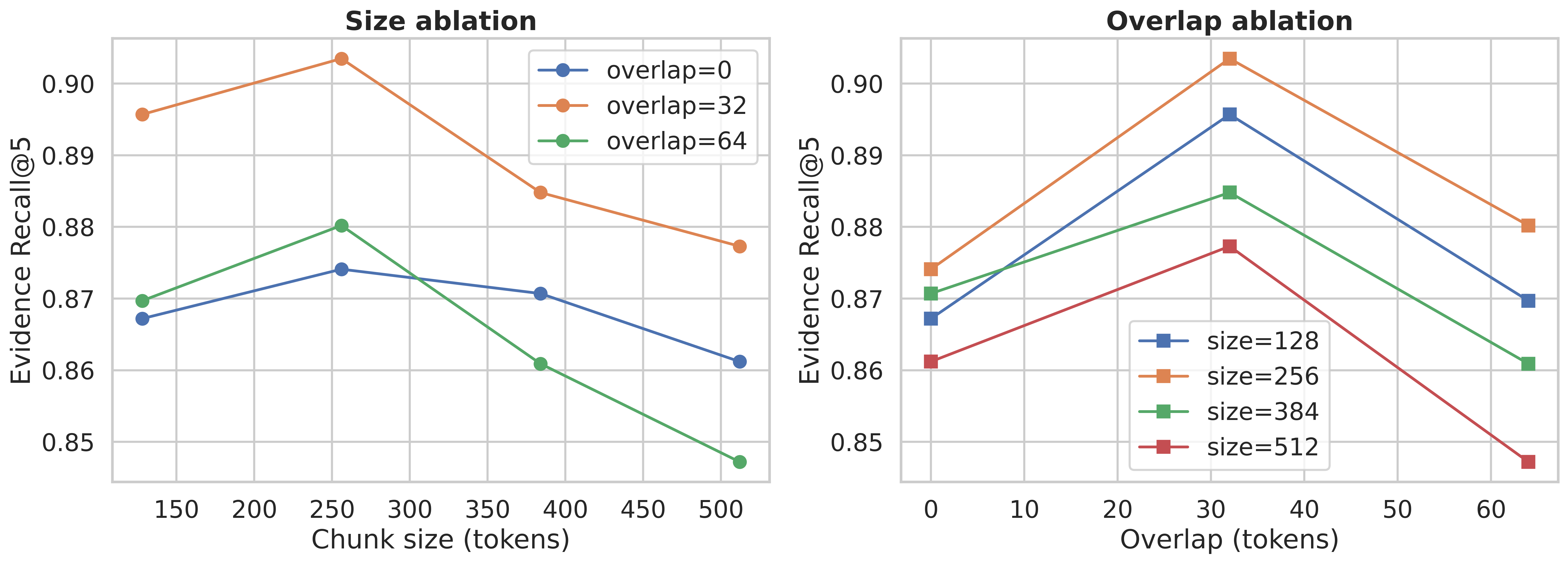}
\caption{Chunk size / overlap ablation, Evidence Recall@5.}
\label{fig:chunk_ablation}
\end{figure}

\paragraph{Chunk size and overlap.} Fixing the recursive chunker, \texttt{BGE-base}, and the \texttt{context\_aware} parser, we sweep chunk size $\in\{128,256,384,512\}$ and overlap $\in\{0,32,64\}$ and report \textbf{Evidence Recall@5} (Figure~\ref{fig:chunk_ablation}). Overlap of $32$ tokens dominates both $0$ and $64$ at every chunk size tested; the best combination overall is $256$-token chunks with $32$-token overlap (Evidence Recall@5 $\approx 0.90$), noticeably ahead of the same size with no overlap ($\approx 0.87$). Larger chunk sizes ($384$, $512$) do not improve over $256$ and mildly underperform it at matched overlap, suggesting $256$ tokens is close to a local optimum for this corpus rather than overlap simply substituting for size.

\begin{figure}[H]
\centering
\includegraphics[width=0.95\textwidth]{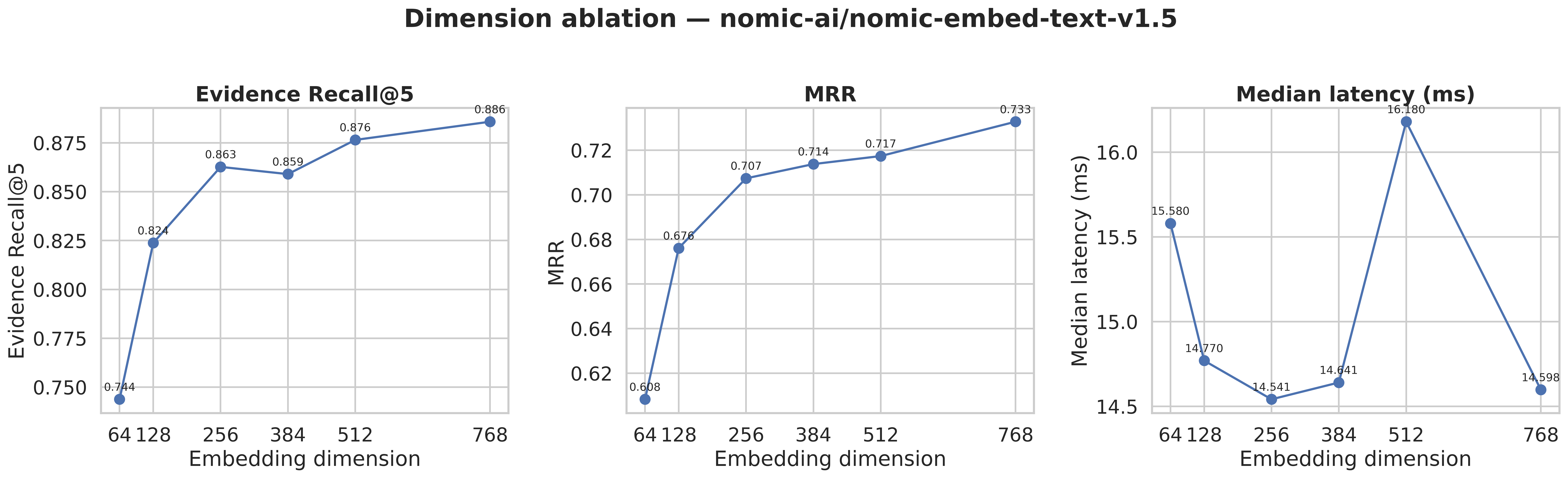}
\caption{Embedding dimension vs.\ Evidence Recall@5, MRR, and median query latency, for the Matryoshka-truncatable \texttt{nomic-embed-text-v1.5} model under the pre-declared \texttt{normal}/\texttt{advanced} parser/chunker selection.}
\label{fig:dim_ablation}
\end{figure}

\paragraph{Embedding dimensionality.} Using a Matryoshka-truncatable embedding model (\texttt{nomic-embed-text-v1.5}) under the pre-declared parser/chunker selection, Figure~\ref{fig:dim_ablation} reports \textbf{Evidence Recall@5}, \textbf{MRR}, and \textbf{median query latency} as a function of truncated embedding dimension. Evidence Recall@5 increases overall from $0.744$ at $64$ dimensions to $0.886$ at the full $768$ dimensions, with most of the gain realized by $256$ dimensions ($0.863$); performance shows a small dip at $384$ dimensions ($0.859$) before recovering at $512$ ($0.876$) and $768$ dimensions ($0.886$). MRR shows the same pattern, rising from $0.608$ at $64$ dimensions to $0.733$ at $768$. Median query latency is comparatively flat across dimensions ($14.5$--$16.2$\,ms), indicating that, for this retriever, the measured per-query latency changes little with dimensionality, suggesting that the main deployment trade-off may instead arise from increased index-storage requirements -- a trade-off practitioners can exploit by truncating well below the full $768$ dimensions with only a modest quality cost.

\begin{figure}[H]
\centering
\includegraphics[width=0.62\textwidth]{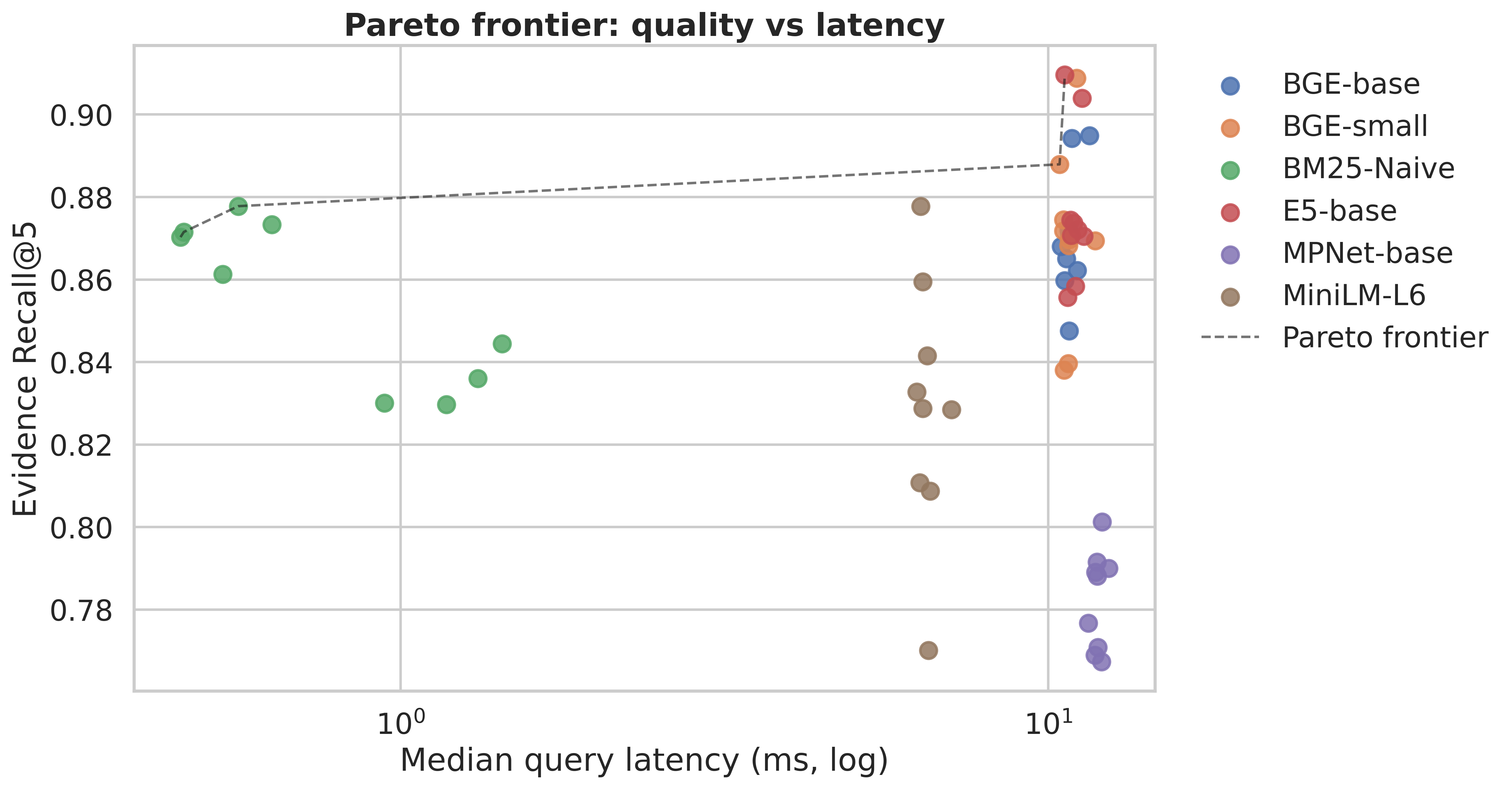}
\caption{Quality/latency Pareto frontier across all 54 configurations.}
\label{fig:pareto}
\end{figure}

\paragraph{Efficiency / quality Pareto frontier.} Figure~\ref{fig:pareto} plots median query latency against Evidence Recall@5 for all $54$ configurations. BM25 configurations dominate the low-latency region (roughly 0.6--1.7\,ms median latency) at only a modest recall cost relative to the best dense configurations (roughly 11\,ms latency), while \texttt{docling}+\texttt{advanced} with \texttt{BGE-small} or \texttt{E5-base} anchor the high-quality end of the frontier. No dense configuration achieves both the highest recall \emph{and} the lowest latency; the choice is a genuine trade-off, not a dominated alternative.

\section{Discussion}

Our results speak most directly to practitioners assembling a RAG pipeline over structured regulatory or policy documents. Three implications stand out. First, \textbf{the interaction between parser and chunker matters more than either component's individual ranking}: the \texttt{context\_aware} parser's heading-injection, which is a common and intuitively-motivated design pattern, does not improve dense retrieval in our corpus and in fact under-performs simpler extraction once combined with structure-aware (\texttt{advanced}) chunking -- most likely because the \texttt{advanced} chunker already performs its own independent section detection, making the parser-level heading tag partially redundant while adding token overhead that a fixed 256-token budget must then accommodate by truncating body text. Second, \textbf{embedding model selection should be validated per-document, not assumed from an aggregate benchmark leaderboard}; a model that performs competitively in general-purpose semantic-similarity benchmarks (MPNet-base) is a significant, specific liability on table-derived content in our regulatory-document setting. Third, \textbf{the near-saturated representation ceiling we observe (\S\ref{sec:ceiling}) is itself informative}: for this class of document, our tested parsing and chunking configurations preserve information well; the remaining bottleneck is ranking, which is exactly where the parser $\times$ chunker $\times$ embedding interaction we measure has practical consequence.

\section{Limitations}
\label{sec:limitations}

\begin{itemize}
    \item \textbf{Four documents, single domain.} All four documents are Indian central-government administrative/legal texts. While we deliberately chose structurally heterogeneous documents within this domain, our findings should not be assumed to transfer to other domains (e.g.\ scientific literature, customer-support knowledge bases) or languages without further validation.
    \item \textbf{LLM-generated questions.} Question--evidence pairs were generated by an LLM under a fixed, verbatim-evidence-constrained schema, with a 10\% manual spot-check and de-duplication pass by the authors, rather than being fully independently human-authored or subjected to formal inter-annotator agreement measurement. We report this plainly rather than presenting the benchmark as a fully human-curated gold standard.
    \item \textbf{No generation-stage evaluation.} This study is retrieval-only. We do not measure whether improved retrieval quality translates into better final answers from a downstream generator. We leave this to future work; a natural extension would add a downstream generation stage.
    \item \textbf{Near-ceiling evidence preservation limits discriminative power for parser/chunker comparisons specifically at the corpus-coverage level} (\S\ref{sec:ceiling}); our reported parser/chunker effects should be interpreted as effects on \emph{ranking}, not on what content survives ingestion, which we consider a finding in itself, not merely a caveat.
    \item \textbf{No cross-encoder reranking stage or hybrid (sparse+dense) baseline} was evaluated; given that no single family dominates across documents, a lightweight hybrid or reranking stage is a natural next comparison point.
\end{itemize}

\section{Conclusion}

We present a controlled, statistically-grounded factorial evaluation of parser, chunking, and embedding choices for retrieval-augmented generation over structurally diverse Indian government regulatory documents. Across an $800$-question, four-document benchmark and $72{,}000$ query-level evaluations, we find that no single retriever family or parser/chunker pairing dominates uniformly; that a popular general-purpose embedding model carries a specific, reproducible weakness on tabular content; and that our corpus's near-saturated evidence-preservation ceiling isolates the retrieval-quality differences we measure to ranking behavior rather than information loss. We release our evaluation harness, corpus manifest, and benchmark to support further work on structure-aware RAG evaluation.

\bibliographystyle{plainnat}

\appendix
\section{Reproducibility and Code Availability}
The complete implementation is publicly available at \url{https://github.com/shubham-61291/pre_generation_rag_retrieval_benchmark}. The file \texttt{pre\_generation\_rag\_retrieval\_benchmark.py} contains all code for corpus construction, question--evidence generation, the full parser $\times$ chunker $\times$ embedding factorial sweep, statistical modeling, and figure generation. Configuration hash, checkpointed run cache, and full per-query result tables (\texttt{master\_results.csv}, $72{,}001$ lines including the header ($72{,}000$ data rows)) are included in the supplementary release. Package versions: \texttt{sentence-transformers}, \texttt{faiss-cpu}, \texttt{rank\_bm25}, \texttt{statsmodels}, \texttt{docling}, \texttt{pypdf} (exact pinned versions in \texttt{requirements.txt}, released alongside the code). All embedding models are used with their published query/passage prompt prefixes where applicable (BGE, E5) and without any prefix otherwise (MiniLM, MPNet). All experiments can be reproduced by running this script with Python 3.9 and the listed dependencies.

\section{Additional Figures}
\label{sec:additional_figures}
This appendix collects supplementary figures from the full evaluation sweep that complement the main-text analyses in \S5.

\begin{figure}[H]
\centering
\includegraphics[width=0.62\textwidth]{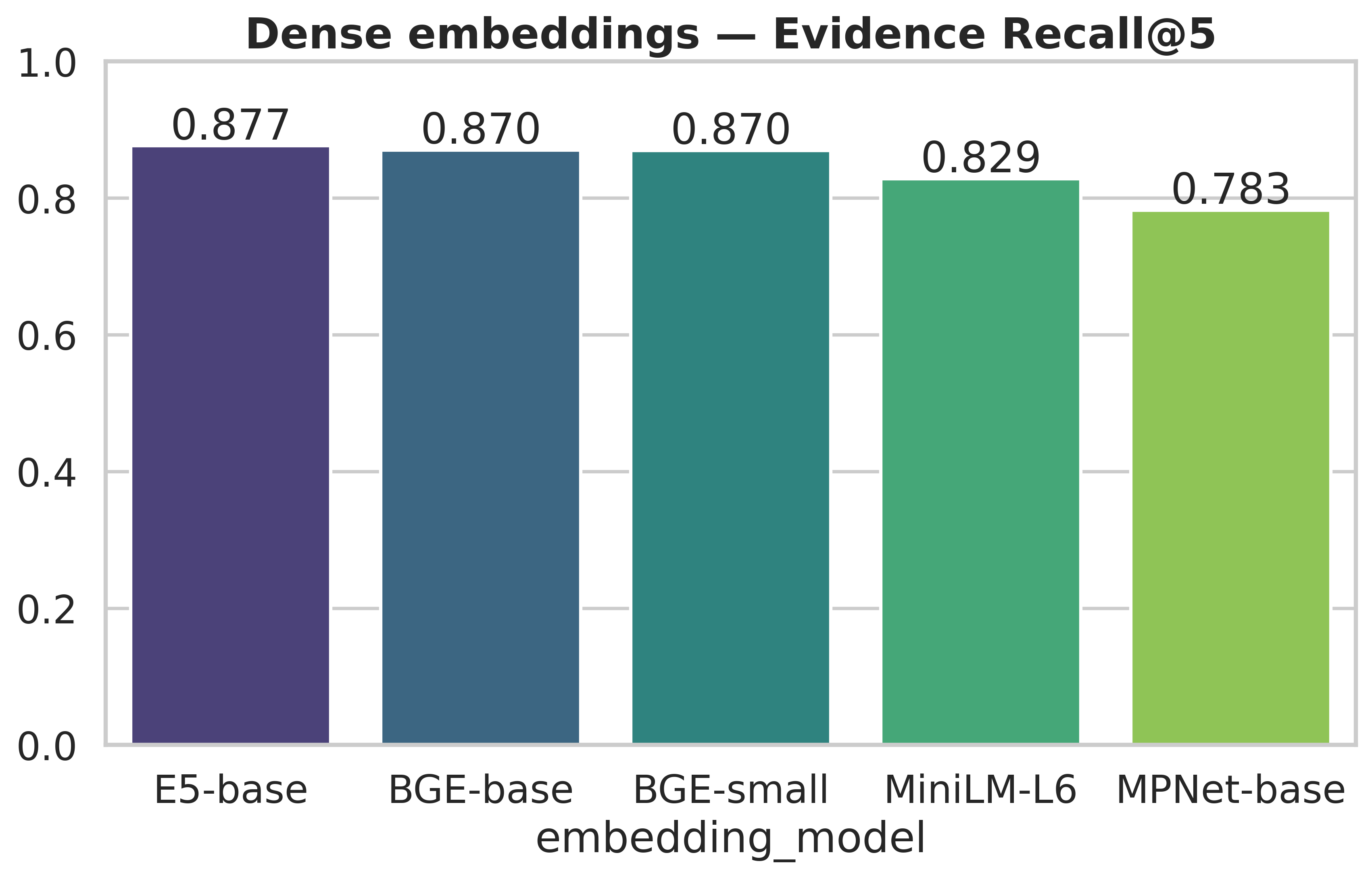}
\caption{Dense embedding models compared directly on Evidence Recall@5, averaged over all parsers, chunkers, and documents. Complements Table~\ref{tab:densevsbm25}.}
\label{fig:dense_embeddings}
\end{figure}

\begin{figure}[H]
\centering
\includegraphics[width=0.55\textwidth]{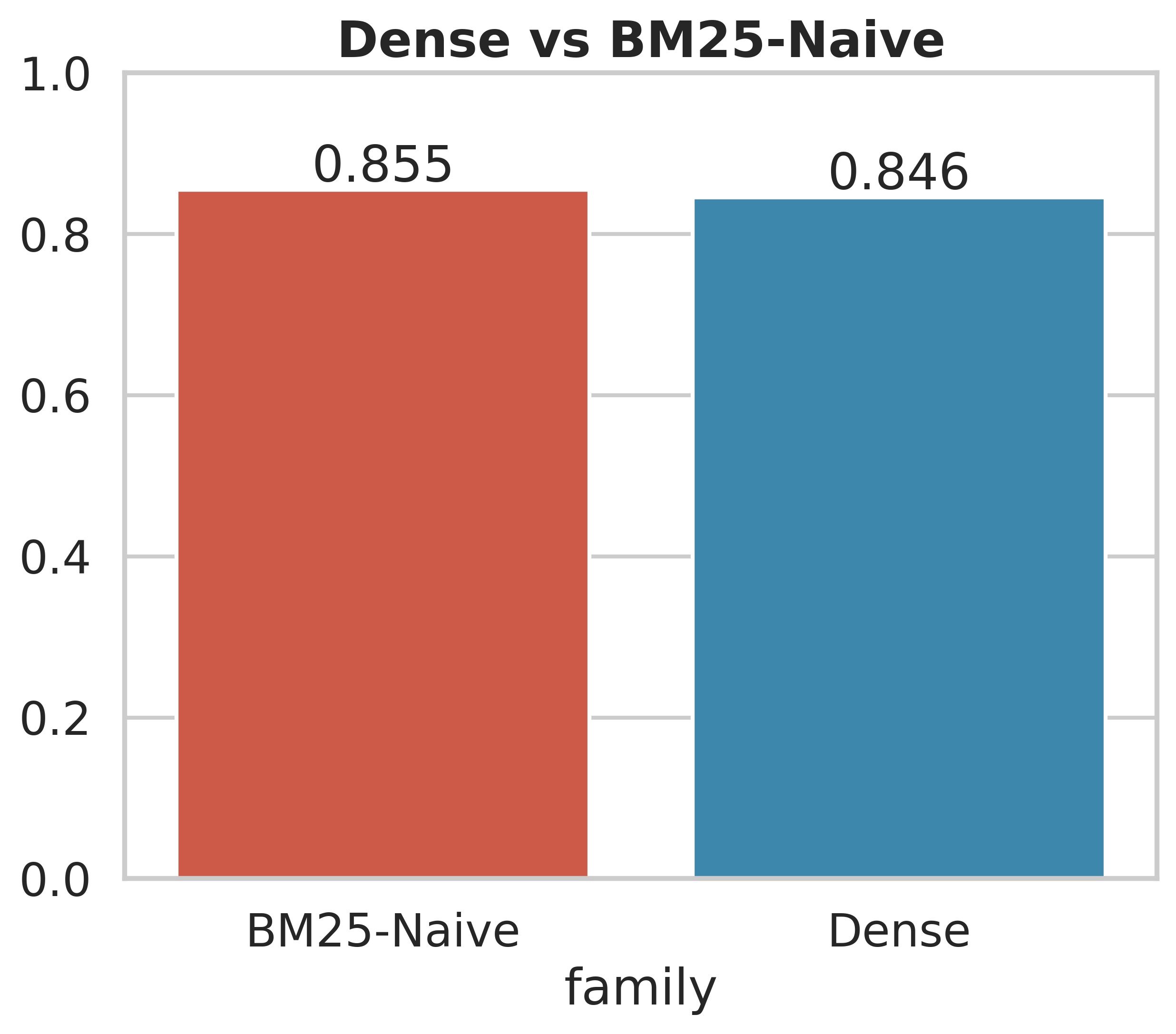}
\caption{Aggregate Evidence Recall@5, dense (best-of-5) vs.\ BM25-Naive, pooled across all four documents. See Figure~\ref{fig:dense_vs_bm25} for the per-document breakdown that motivates RQ2.}
\label{fig:dense_vs_bm25_agg}
\end{figure}

\begin{figure}[H]
\centering
\includegraphics[width=0.7\textwidth]{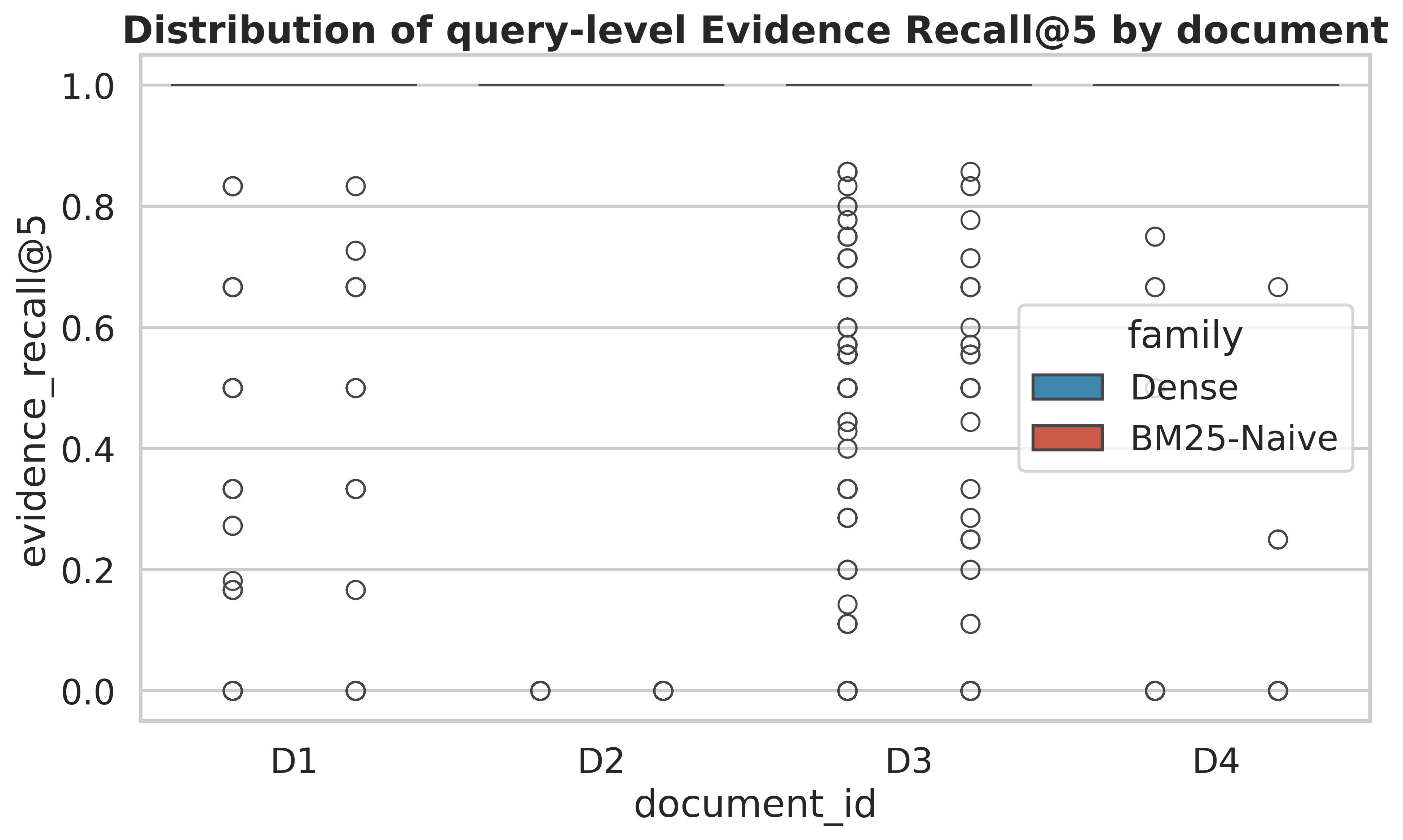}
\caption{Distribution of query-level Evidence Recall@5 by document and retriever family. The wide spread on D3 relative to D1/D2/D4 illustrates why document-level means (Figure~\ref{fig:dense_vs_bm25}) can mask substantial per-query variance within a single document.}
\label{fig:per_document_dist}
\end{figure}

\begin{figure}[H]
\centering
\includegraphics[width=0.85\textwidth]{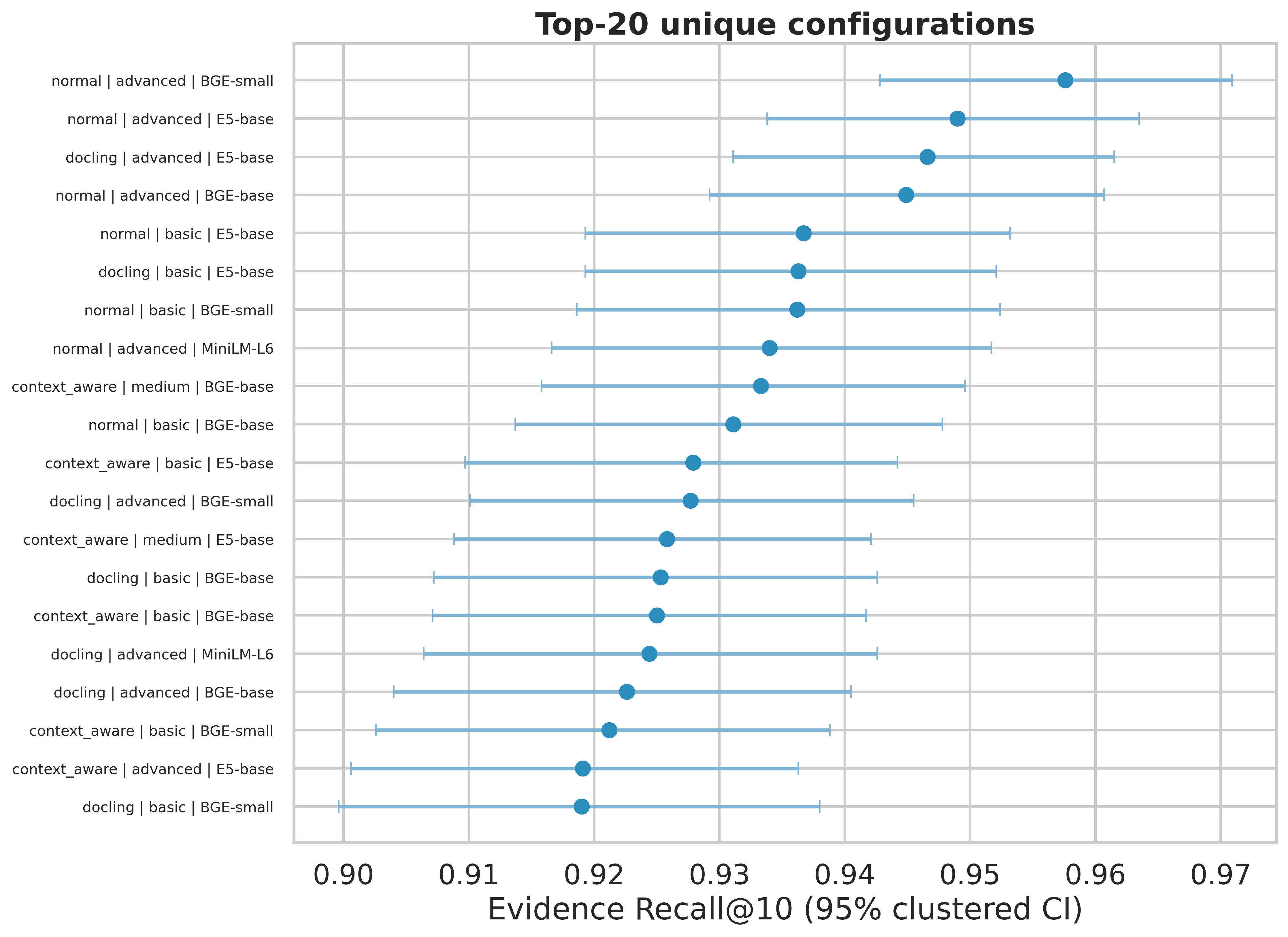}
\caption{Top-20 unique (parser, chunker, embedding) configurations ranked by Evidence Recall@10, with 95\% clustered bootstrap confidence intervals (2{,}000 resamples over query-level means). Overlapping intervals among the top configurations indicate that several parser/chunker/embedding combinations are statistically indistinguishable at the top of the leaderboard.}
\label{fig:bootstrap_forest}
\end{figure}

\begin{figure}[H]
\centering
\includegraphics[width=0.62\textwidth]{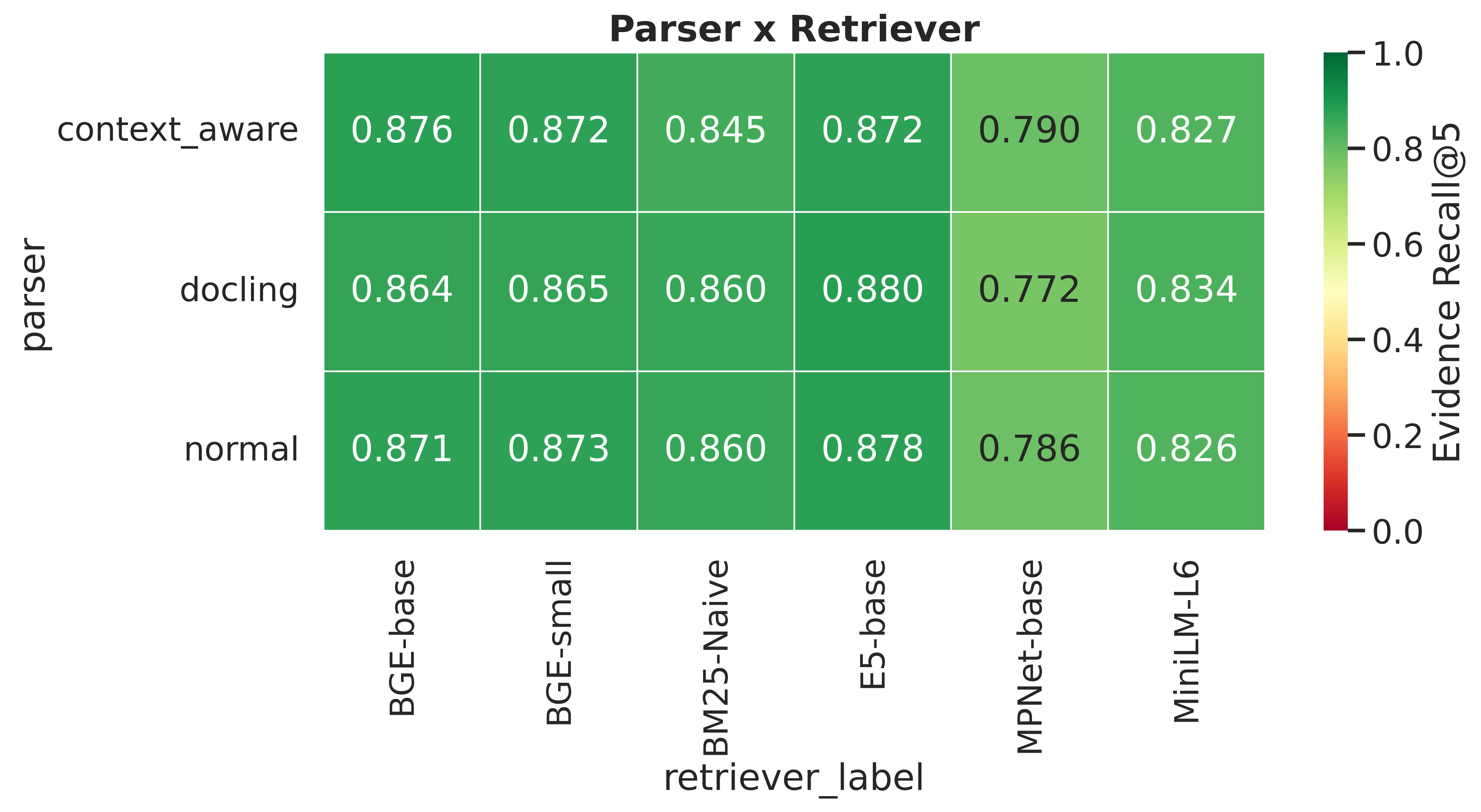}
\caption{Mean Evidence Recall@5 by parser and retriever, averaged over all three chunkers. MPNet-base's underperformance (\S\ref{sec:densevsbm25}) is consistent across all three parsers rather than isolated to one.}
\label{fig:parser_x_retriever}
\end{figure}

\begin{figure}[H]
\centering
\includegraphics[width=0.72\textwidth]{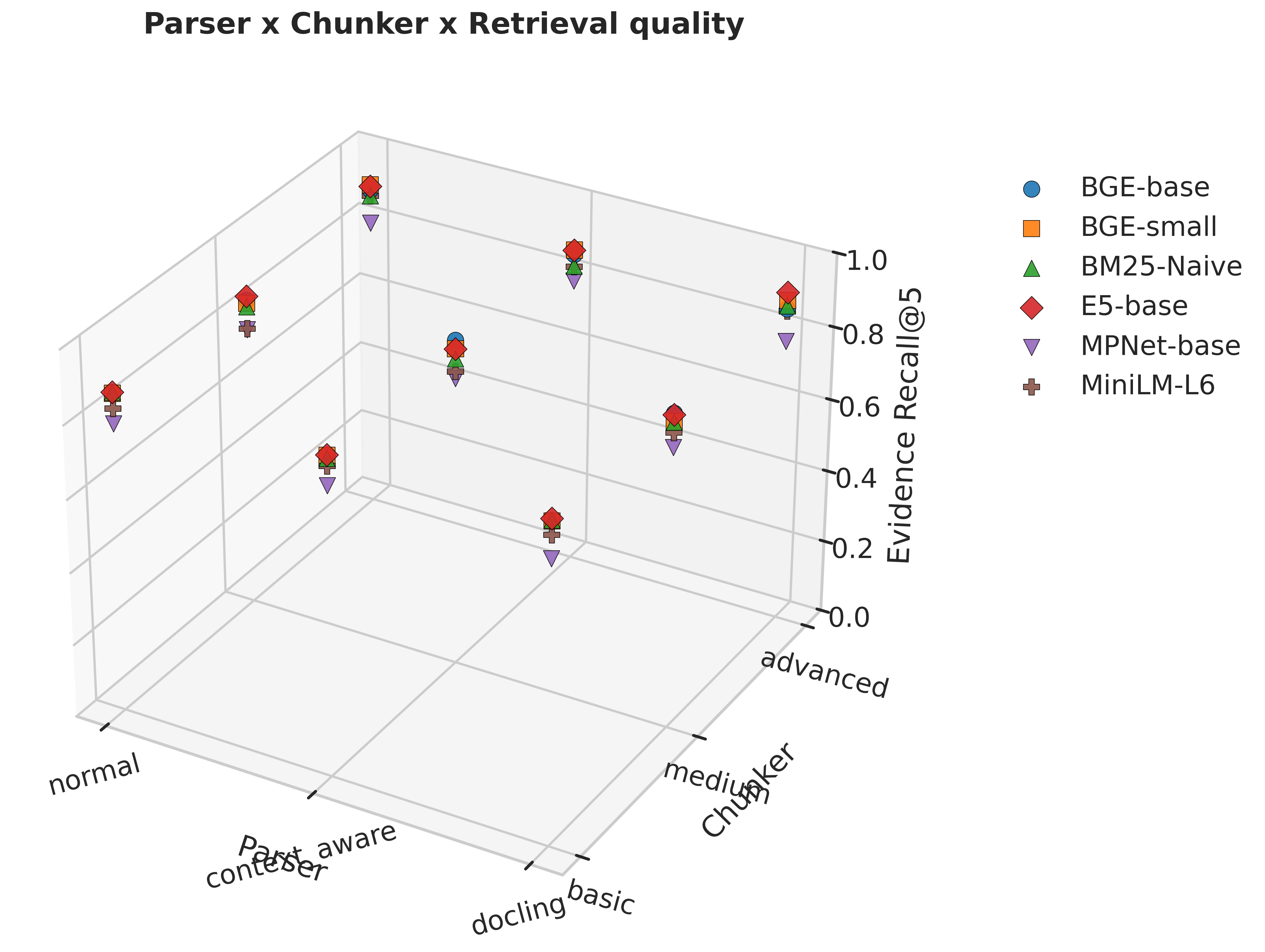}
\caption{Evidence Recall@5 as a joint function of parser and chunker, faceted by embedding model. A three-dimensional view of the interaction quantified in Table~\ref{tab:mixedlm}.}
\label{fig:3d_scatter}
\end{figure}

\begin{figure}[H]
\centering
\includegraphics[width=0.72\textwidth]{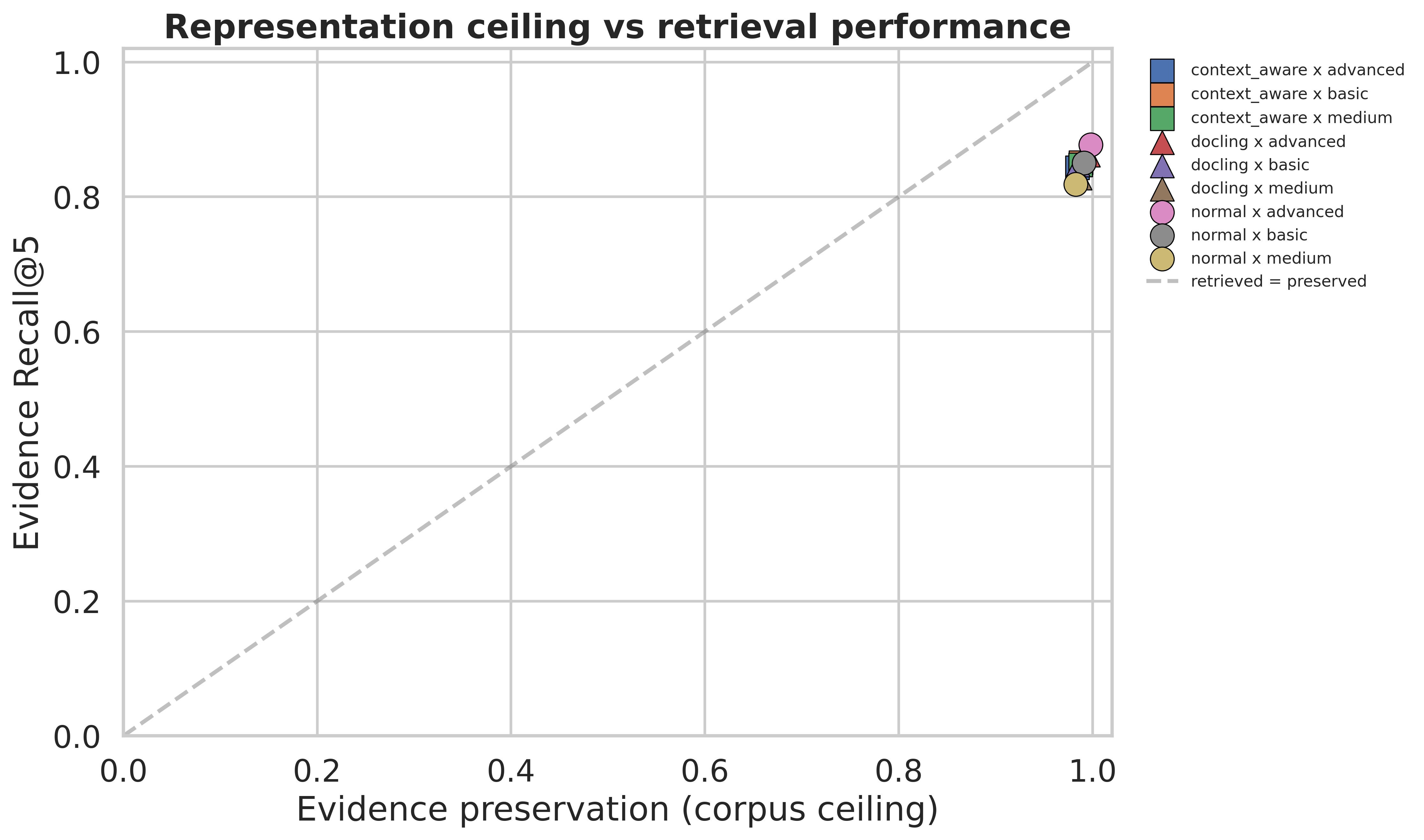}
\caption{Evidence Recall@5 vs.\ corpus-level evidence preservation (the representation ceiling of Table~\ref{tab:ceiling}) for every parser $\times$ chunker combination. All points sit just below the retrieved-equals-preserved diagonal and close to the $x=1$ boundary, visually confirming RQ3: retrieval quality is bottlenecked by ranking, not by information loss during ingestion.}
\label{fig:representation_ceiling}
\end{figure}

\begin{figure}[H]
\centering
\includegraphics[width=0.62\textwidth]{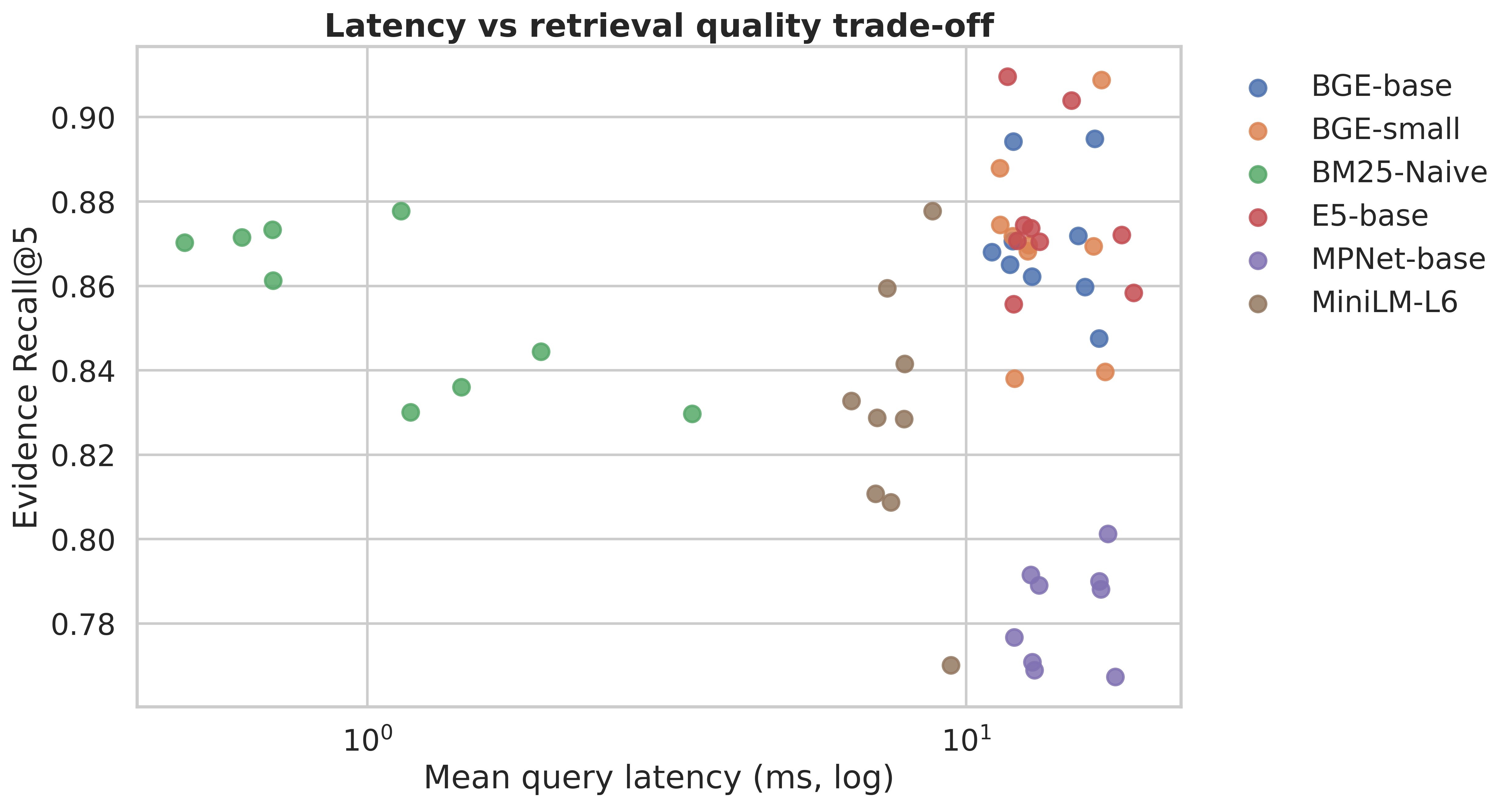}
\caption{Per-configuration mean query latency vs.\ Evidence Recall@5, colored by retriever. An alternative view of the latency/quality relationship underlying the Pareto frontier in Figure~\ref{fig:pareto}.}
\label{fig:latency_vs_quality}
\end{figure}

\begin{figure}[H]
\centering
\includegraphics[width=0.85\textwidth]{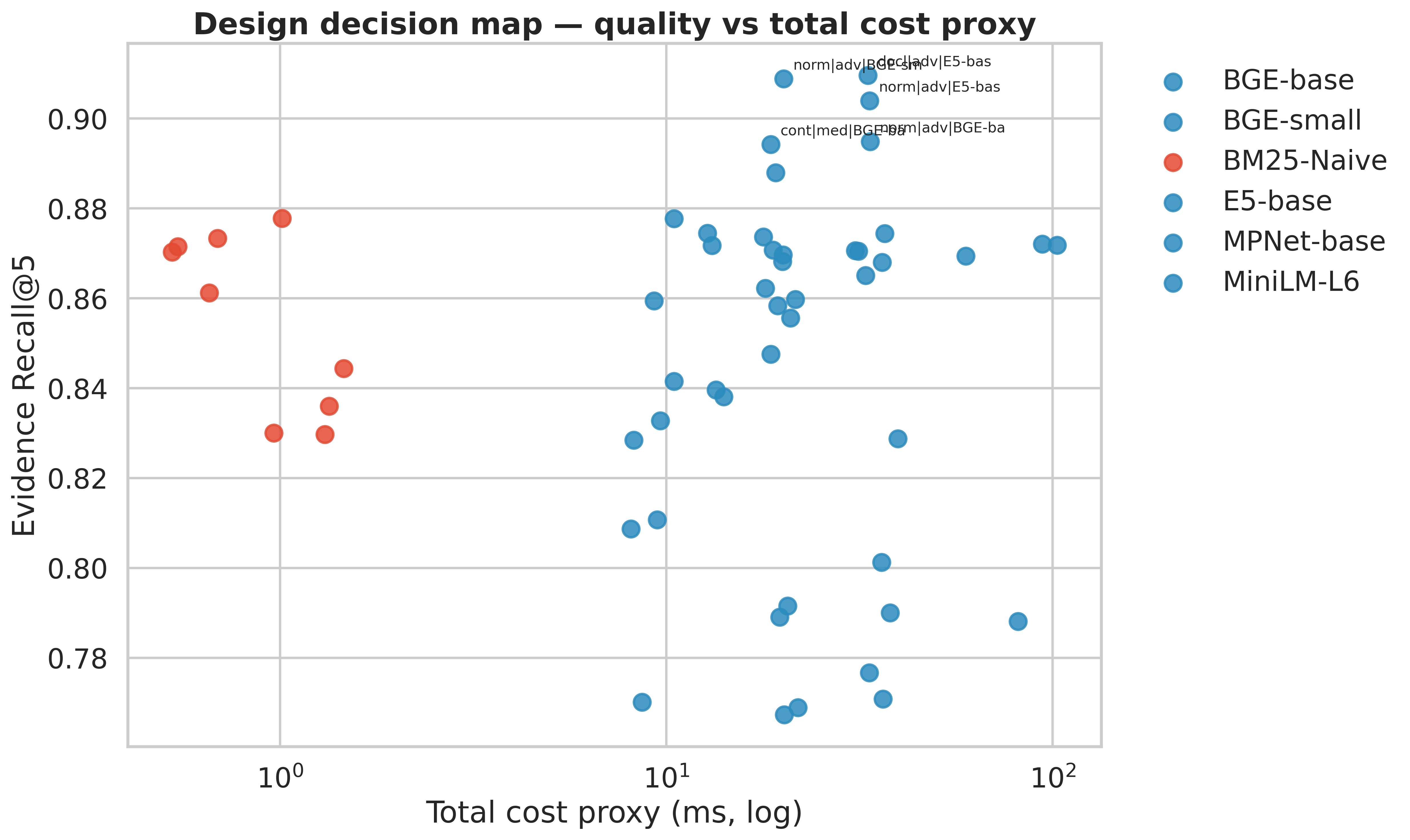}
\caption{Design decision map: Evidence Recall@5 vs.\ a total-cost proxy (log-scale latency) for all 54 configurations, with the top five dense configurations labeled. BM25-Naive configurations cluster at the low-cost end; the labeled dense configurations mark the highest-quality region of the trade-off space.}
\label{fig:design_decision_map}
\end{figure}

\end{document}